\documentclass[runningheads]{llncs}
\usepackage{graphicx}
\usepackage{xcolor}
\usepackage{amsmath}
\usepackage{amsfonts}
\usepackage{wrapfig}
\usepackage{multirow}

\usepackage{amsfonts}
\usepackage{amsmath,amssymb}

\usepackage{bbding}
\usepackage[misc]{ifsym}

\usepackage[pagebackref=false,breaklinks=true,colorlinks,bookmarks=false]{hyperref}

\begin{document}
\title{Learning Multi-resolution Graph Edge Embedding for Discovering Brain Network Dysfunction in Neurological Disorders \vspace{-5pt}}
\titlerunning{MENET}
%


\author{Xin Ma\inst{1}$^\text{(\Letter)}$ \and
Guorong Wu\inst{2} \and
Seong Jae Hwang\inst{3} \and 
Won Hwa Kim\inst{1,4}
\vspace{-3pt}
}


%
\authorrunning{Xin Ma et al.}
%


\institute{University of Texas at Arlington, Arlington, USA \\ \email{xin.ma@mavs.uta.edu} \\ \and
University of North Carolina at Chapel Hill, Chapel Hill, USA\\ \and 
University of Pittsburgh, Pittsburgh, USA \\ \and
POSTECH, Pohang-si, South Korea
\vspace{-15pt}
}

\maketitle              

\begin{abstract}
Tremendous recent literature show that associations between different brain regions, i.e., brain connectivity, provide early symptoms of neurological disorders. Despite significant efforts made for graph neural network (GNN) techniques, their focus on graph nodes makes the state-of-the-art GNN methods not suitable for classifying brain connectivity as graphs where the objective is to characterize disease-relevant network dysfunction patterns on graph links. To address this issue, we propose Multi-resolution Edge Network (MENET) to detect disease-specific connectomic benchmarks with high discrimination power across diagnostic categories. The core of MENET is a novel graph edge-wise transform that we propose, which allows us to capture multi-resolution ``connectomic'' features. Using a rich set of the connectomic features, we devise a graph learning framework to jointly select discriminative edges and assign diagnostic labels for graphs. Experiments on two real datasets show that MENET accurately predicts diagnostic labels and identify brain connectivities highly associated with neurological disorders such as Alzheimer's Disease and Attention-Deficit/Hyperactivity Disorder.

\end{abstract}
\vspace{-20pt}
\section{Introduction}
\label{sec:intro}
\vspace{-8pt}

Many neuroimaging studies operate with data from a population of cohort that can be stratified into two or more groups (e.g., diseased vs. control). Given registered imaging measures acquired from participants, 
contrasting the different groups at each pixel/voxel over the whole brain identifies those regions that are affected by the variable of interest (e.g., disease or risk factors) \cite{friston2003statistical}. 
However, due to poor correlation between cognitive changes and pathological features from images, 
recent studies motivate that characterizing changes in the {\em brain connectivity} or {\em network} that comprise several affected regions yield a better understanding of the brain over traditional spatial analyses \cite{greicius2003functional,bullmore2009complex,ma2020enriching}. 

In brain connectivity studies, an individual brain is divided into registered regions of interests (ROIs) 
and associations across the ROIs are defined via functional/structural images, 
formulated as a ``graph'' that consists of nodes and edges. 
Changes in individual connectivity have shown strong implication of 
neurological diseases with evidence 
such as Parkinson's \cite{ng2017distinct} and Alzheimer's disease (AD) \cite{dennis2014functional}.
Still, few frameworks exist for traditional prediction tasks for 
registered but topologically variant graphs {\em without node-specific features}.

Existing techniques including Graph Neural Network (GNN) methods 
mainly  focus on analyzing features defined on graph nodes, 
where the graph is given as the domain of the signal \cite{kipf2016semi,velivckovic2017graph}.
GNNs for point clouds may work  
but their nodes are merely arbitrarily sampled points over surfaces of non-rigid shapes \cite{landrieu2018large}. 
Notice that such settings are not in the interest of this study 
where we deal with alternations in the ``connectivity'', i.e., edges, instead of measures at each node. 
Graph kernels that compare local substructures \cite{riesen2009graph,shervashidze2009efficient} 
are often inadequate for training, 
and alternative deep learning methods for graph such as \cite{zhang2018end,simonovsky2017dynamic} 
lose interpretability in space particularly with dense graphs. 
For brain connectivity analysis, the method must be able to investigate local 
edge-wise variation and predict global diagnostic labels to identify disease-specific symptoms.
Hence, it is critical to develop a framework 
that can {\em adaptively} transform graph ``edges'' with graphs' inherent {\em structure} of the graphs to increase {\em sensitivity} but maintain location-wise {\em interpretability} even with small sample sizes. 

\begin{wrapfigure}{r}{0.5\textwidth}
\vspace{-2.2em}
\centering
\includegraphics[width=.12\textwidth, height = 1.62cm]{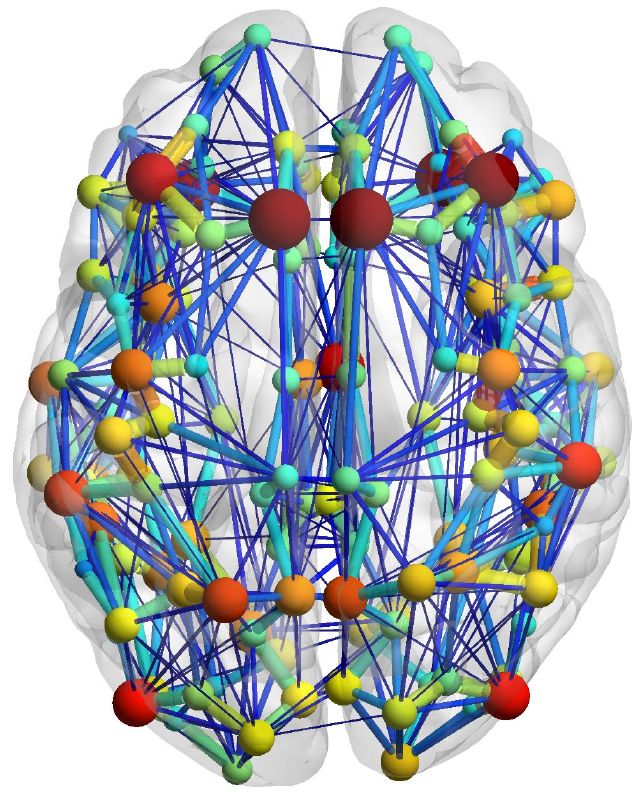}\hspace{-0.3em}
\includegraphics[width=.12\textwidth, height = 1.62cm]{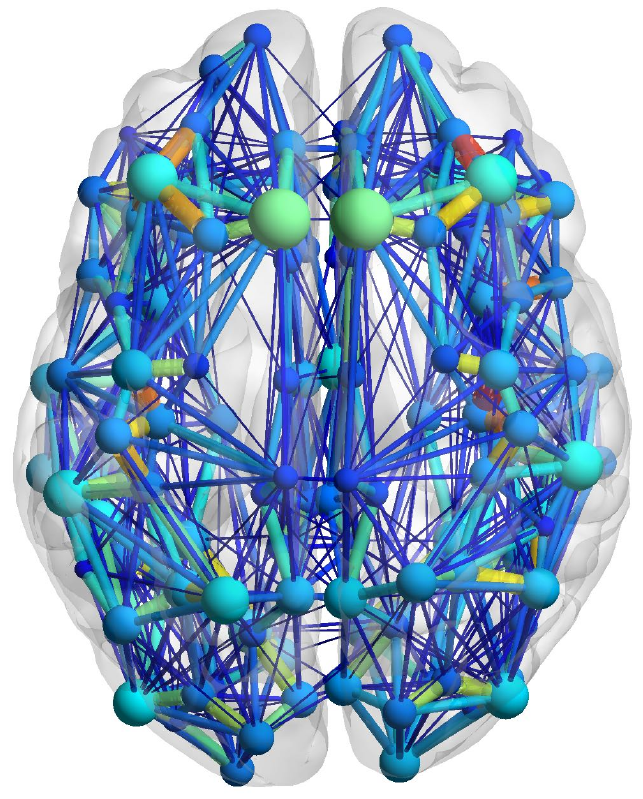}\hspace{-0.3em}
\includegraphics[width=.12\textwidth, height = 1.62cm]{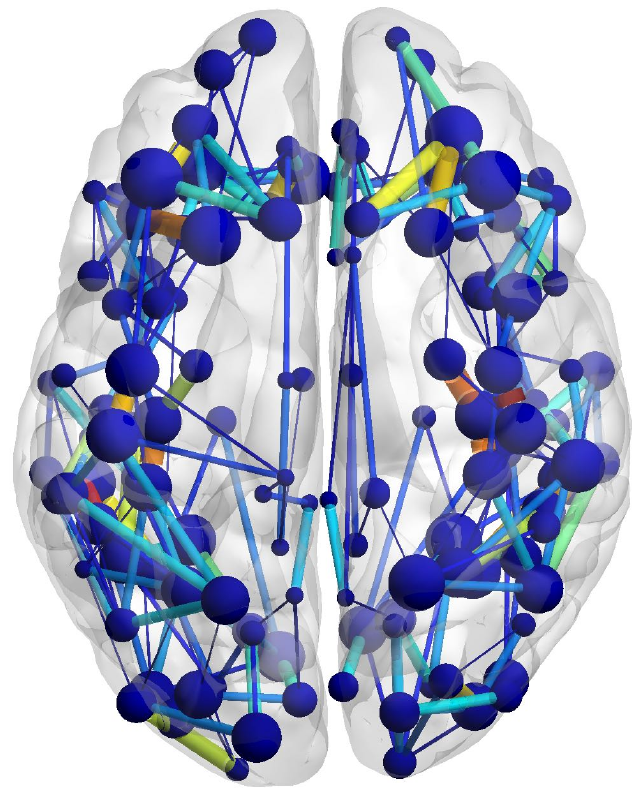}\hspace{-0.3em}
\includegraphics[width=.12\textwidth, height = 1.62cm]{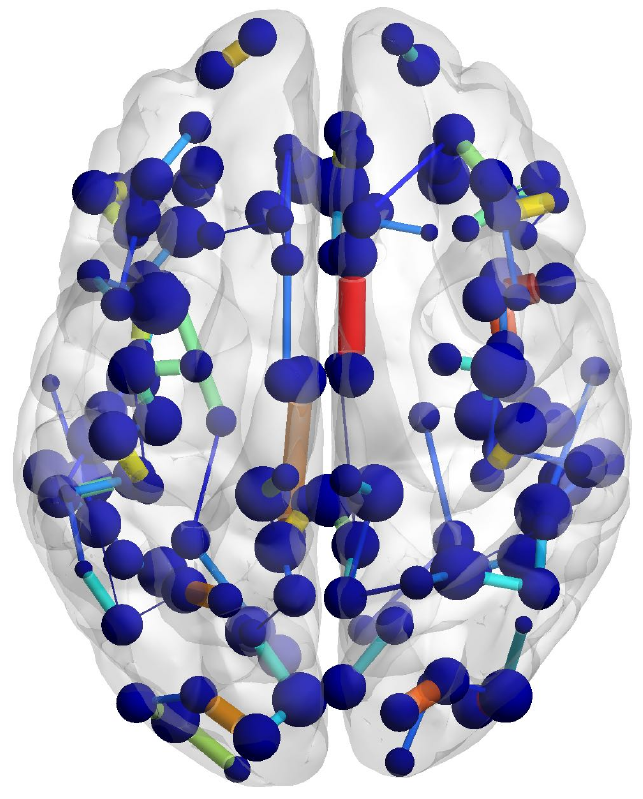}
\vspace{-7pt}
\caption{\scriptsize Example of multi-resolution representation of brain connectivity. 
1) original network, 2)-4) filtered network at $s= 0.3, 0.8, 1.8$. 
Sparsity of the network increases and nodal degree decreases (red to blue) as the scale changes.}
\label{fig:with_gr_2g}
\vspace{-2.3em}
\end{wrapfigure}
To achieve the objectives above, 
we propose a new convolution neural network (CNN) framework for graph edges 
that derives flexible multi-resolution features 
sensitive to topological variations in graphs (example shown in Fig.~\ref{fig:with_gr_2g}). 
The key 
of our method is 
at the adaptive graph edge transform 
--- adopting ideas from spectral graph wavelet transform \cite{Hammond2011129}, 
we define a novel {\em multi-resolution edge transform} 
that deals with a positive semi-definite (p.s.d.) matrix. 
Defining a graph specific orthonormal tensor as a surrogate to transform a p.s.d. matrix, 
we obtain its multi-resolution representation 
with trainable kernel functions in a dual space.  
Such an {\em edge-wise} analysis is 
at the core of brain connectivity analysis identifying which of the brain connectivities are 
significantly related to disease-specific variables.
The  {\bf contributions} of our work are, 
	1) defining a novel graph edge transform that derives flexible representation of graph edges, 
	2) proposing a framework that can efficiently train and classify diagnostic labels for graphs 
	as well as identify disease-specific brain connectivity, 
	3) extensive empirical results on two independent real brain connectivity data for AD 
	and Attention-Deficit/Hyperactivity Disorder (ADHD) to validate our framework. 
Our discoveries identifying subtle variations in brain connectivity
align with other on-going studies, 
suggesting that 
our framework has potential for various brain disorder analyses.  
\vspace{-10pt}
\vspace{-5pt}
\subsection{Related Work}
\label{sec:related}
\vspace{-5pt}
{\bf Graph Kernels (GK):} GK methods map graphs into a Hilbert space, which can be fed to downstream classifiers (e.g., Support Vector Machine (SVM)) to perform graph classification \cite{gartner2003graph}. 
The designs of kernels include neighborhood aggregation (e.g., Weisfeiler-Lehman \cite{shervashidze2011weisfeiler}), extraction of subgraph patterns (e.g., Graphlet \cite{shervashidze2009efficient}, and walks/paths (e.g., shortest-path \cite{borgwardt2005shortest} and random walk~\cite{kang2012fast}). 

\noindent{\bf GNN:} 
Graph Convolution Network (GCN) methods transform ``node features'' into informative node embeddings by aggregating and propagating node features 
via graph convolution.
Such node embeddings have shown to excel at node classification \cite{kipf2016semi} 
and graph classifications \cite{velivckovic2017graph} 
with specific pooling mechanisms, e.g.,  
Dynamic  Graph  CNN (DGCNN), one of the state-of-the-art methods, uses a specialized SortPooling layer
for graph classification \cite{dgcnn}. 
Despite their success, GCNs and their variants rely heavily on the node features provided by the data which, in fact, may not even be provided in certain datasets.
In other words, graphs may \textit{only} come with edge features, leading to sub-optimal use of GCN by reluctantly utilizing weak node features such as nodal degrees.  

\noindent{\bf Benefits of our work.} Unlike the two approaches above, we use 
parametric kernels can adaptively derive beneficial representations for graph classification 
that can be directly applied to edge measures instead of node features.
Different from the methods 
that define multi-resolution 
in the native graph space 
with hops on nodes 
\cite{xu2019mr} or 
thresholds on edges~\cite{jie2014topological}, we  use a dual space and smooth kernels that can be described by the theory of traditional wavelet transform. 
\vspace{-10pt}
\section{Proposed Method}
\vspace{-5pt}
The key in our approach is to derive the multi-resolution representation of \textit{edges} by developing a novel graph edge transform, which we introduce below. 

\vspace{-8pt}
\subsection{Multi-resolution Graph Edge Transform}
\label{sec:multi-resol-graph}
\vspace{-5pt}
Consider a set of graphs with registered nodes with label $y$ assigned for each graph $G$. Each graph with $N$ nodes is represented as its adjacency matrix $A_{N \times N}$, whose non-negative element $a_{ij}$ denotes edge weight (i.e., measure of association) 
between the $i$-th and $j$-th nodes. 
Given an adjacency matrix $A$ (e.g., a brain connectivity), 
one can 
derive a degree matrix $D_{N \times N}$ as a diagonal matrix whose $i$-th diagonal is the sum of edge weights 
connected to the $i$-th node (i.e., volume). 
Its graph Laplacian $\mathcal{L}$ is defined as $D$$-$$A$, 
and a normalized one 
is defined as $\mathcal L_{norm}$$=$$D^{-1/2}\mathcal{L}D^{-1/2}$. 
Both Laplacians, $\mathcal{L}$ and $\mathcal L_{norm}$, are symmetric and p.s.d. with 
non-negative eigenvalues $\lambda_{\ell}$ 
and orthonormal eigenvectors $u_{\ell}$. 
Previous works 
used the $\lambda_\ell$ and $u_\ell$ to define 
a wavelet transform on node signals 
and 
constructed 
GCNs for signals defined on the nodes $n$, i.e., $f(n)$ \cite{Hammond2011129,kipf2016semi}. 
However, notice that $f(n)$ is not of our interest but 
we define a transform for $f(e)$, i.e., weights on the edges $e$, to derive ``multi-resolution'' views. 

Let us first decompose a graph Laplacian $\mathcal{L} = \sum^{N-1}_{l=0} \lambda_l u_l u_l ^T =U\Lambda U^{T}$
where $U = [u_0, u_1, \cdots , u_{N-1} ]$ and $\Lambda$ is a diagonal matrix whose $(l+1)$-th diagonal 
is $\lambda_l$. 
Traditional continuous wavelet transform 
suggests that derive multi-resolution representation of signals can be derived
using an orthogonal transform and defining scales in a dual space \cite{mallat1999wavelet,Hammond2011129,kim2016latent}. 
Hence, we define an orthonormal basis for a ``matrix'' using an outer product of $u_l$, together with a kernel $k(\cdot)$ as $\psi_{l,s}(i,j) = k(s\lambda_l) u_l(i) u_l(j)$
and a transform 
using the $\psi_{l,s}(i,j)$ is 
defined as
{\footnotesize
\begin{align}
	\beta_{ \mathcal{L} ,l}(s) &= \langle \mathcal{L}, \psi_{l} \rangle
	= \sum_i \sum_j k(s\lambda_l) u_l(i) u_l(j) \sum_{l'=0} ^{N-1} \lambda_{l'} u_{l'} (i) u_{l'} (j)
	= \lambda_{l} k(s\lambda_l)
\end{align}}which yields a resultant coefficient $\beta_{\mathcal{L}, \ell}(s)$.
Note that its inverse is
	\begin{equation}
	\footnotesize
	\label{eq:L_recon}
	\mathcal{L}(i,j) = \frac{1}{C_k} \int_0 ^\infty \sum_{l=0} ^{N-1} \beta_{\mathcal{L},l}(s) \psi_{s,l}(i,j) \frac{ds}{s} 
	\end{equation}
which
reconstructs $\mathcal{L}$ with a kernel normalization constant $C_k = \int_0^\infty \frac{k(x)^2}{x}dx$. 

\begin{lemma} (Graph Laplacian Admissibility Condition) 
Given a kernel dependent normalization constant $C_k = \int_0^\infty \frac{k(x)^2}{x}dx < \infty$, 
the original graph Laplacian $\mathcal L$ can be perfectly reconstructed via the inverse transformation. 
\begin{proof}
Projecting the coefficients $\beta_{ \mathcal{L} ,l}(s)$ back to the original domain with $\psi_{l,s}$, 
{\footnotesize
	\begin{align}
	&\frac{1}{C_k} \int_0 ^\infty \sum_{l=0}^{N-1} \beta_{\mathcal{L},l}(s) \psi_{l,s} (i,j) \frac{ds}{s} = \frac{1}{C_k} \int_0 ^\infty \sum_{l=0} ^{N-1} \lambda_l k(s\lambda_l)^2 u_l(i) u_l(j) \frac{ds}{s} \nonumber \\
	&~~~~~= \sum_{l=0} ^{N-1} \lambda_l [\frac{1}{C_k} \int_0 ^\infty \frac{k(s\lambda_l)^2}{s}ds] u_l(i) u_l(j) \nonumber 
	= \mathcal{L}(i,j) 	
	\end{align}
}and $\mathcal{L}(i,j)$  is obtained by substituting $s\lambda_l = x$. 
\end{proof}
\end{lemma}

The Lemma above 
follows the traditional admissibility condition in wavelet transform \cite{mallat1999wavelet}, i.e., 
a superposition of multi-resolution representation of $\mathcal{L}$ 
over scales $s$. 
It lets us to define new representation $\mathcal{L}_s$ at different scales $\mathbf{s}$:
	\begin{equation}
	\footnotesize
	\mathcal{L}_s(i,j) = \sum_{l=0} ^{N-1} \lambda_l k(s\lambda_l)^2 u_l(i) u_l(j)
	\label{eq:multi_resol_graph}
	\end{equation}
by focusing at a specific scale $s$. It is given as a matrix operation as
\begin{equation}
\footnotesize
\begin{aligned}
	\mathcal{L}_{s} = U k(s\Lambda)^2\Lambda U^{T} \label{eq:Ls}
\end{aligned}
\end{equation}
where $k(s\Lambda)^2$ is a diagonal matrix with the $k(\cdot)$ applied at each $\lambda$.  
The shape of $k(\cdot)$ determines the shape of wavelet-like basis 
and multi-resolution views of $\mathcal L$. 
For regular images, this yields filtered images (e.g., band-pass filtered signal), and our framework extracts similar representations of 
edges $f(e)$ as in Fig.~\ref{fig:with_gr_2g}.

\vspace{-8pt}
\subsection{Efficient Graph Matrix Transform}
\vspace{-5pt}
The transform \eqref{eq:multi_resol_graph} requires eigendecomposition of a graph Laplacian (or estimating a partial set of eigenvectors) 
which can be computationally burdening, 
especially with a large number of nodes or graph samples. 
We therefore suggest an approximation of \eqref{eq:multi_resol_graph} 
that significantly reduces computation with marginal error, 
which extends the approximation 
in \cite{Hammond2011129} to 
our graph matrix transform. 

For this, let us assume $g(\Lambda) = \Lambda k(s\Lambda)^{2}$ from \eqref{eq:Ls} and $\tilde{\Lambda} = \frac{2}{\lambda_{max}} \Lambda-I_{N}$ since the largest eigenvalue of a normalized graph Laplacian is bounded by 2. 
Then, $\tilde{\Lambda} = \Lambda - I_{N}$ and $g(\tilde{\Lambda}) = g(\Lambda-I_{N})$. 
If we expand $g(\Lambda)$ with $I_{N}$ then,
\begin{equation}
\footnotesize
\begin{aligned}
	g(\Lambda) &= \overset{\infty}{\underset{n=0}{\sum}} \frac{g^{(n)}(I_{N})}{n!}(\Lambda -I_{N})^{n} = \overset{\infty}{\underset{n=0}{\sum}} \frac{g^{(n)}(0_{N})}{n!}\tilde{\Lambda}^{n}. 
\end{aligned}
\end{equation}
Considering that the elements of $\tilde{\Lambda}$ are in [-1, 1], 
we assume there exists a positive $K$ where if $n > K$, $\tilde{\Lambda}^{n} \rightarrow 0$. Here, $g(\Lambda)$ can be approximated as 
{
\begin{equation}
\footnotesize
    g(\Lambda) \approx \overset{K}{\underset{n=0}{\sum}} \frac{g^{(n)}(0_{N})}{n!}\tilde{\Lambda}^{n}.
\end{equation}
}Since $U\Lambda^{m} U^{T} = (U\Lambda U^{T})^{m}$, $\mathcal{L}_{s}$ now can be written as 
\begin{equation}
\footnotesize
	\mathcal{L}_{s} = U g(\Lambda) U^{T} \approx \overset{K}{\underset{n=0}{\sum}} \frac{g^{(n)}(0_{N})}{n!} U \tilde{\Lambda}^{n} U^{T} \approx \overset{K}{\underset{n=0}{\sum}} \frac{g^{(n)}(0_{N})}{n!} \tilde{\mathcal{L}}^{n}
\end{equation}
where $\tilde{\mathcal{L}} = \mathcal{L} - I_{N\times N}$. 
This approximation lets our overall framework 
more practical 
to handle large scale graph data.

\vspace{-10pt}
\subsection{Network Architecture}
\begin{figure*}[t]
	\centering
    \includegraphics[width=0.93\linewidth, height = 3.3cm]{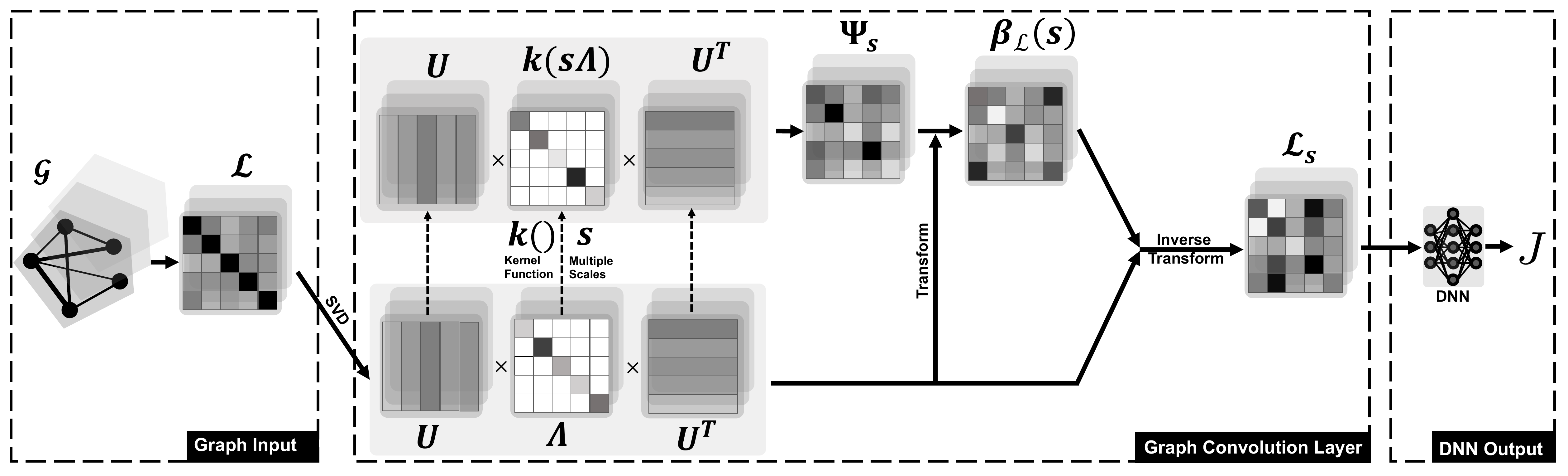}    
    \vspace{-1em}
	\caption{\scriptsize
		Overall architecture of MENET. A graph matrix is transformed to yield multi-resolution representations, and then fully connected (FC) DNN is applied at the end. Error is backpropagated to train the weights $\mathbf{W^h}$
		and update the scales $\mathbf{s}$ to obtain the optimal representations. 
		}
	\label{fig:pipeline}
	\vspace{-2em}
\end{figure*}

\vspace{-5pt}
\label{sec:MGNN}
Based on the 
matrix transform \eqref{eq:multi_resol_graph} as a convolution, 
we propose 
{\em Multi-resolution Edge Network (MENET)}
which is a novel CNN framework that
utilizes multi-resolution representations of graph edges for classification. 
Fig.~\ref{fig:pipeline} illustrates the overall pipeline with the following components: 

{\bf Convolution layer:} Taking a graph as an input, 
it returns multi-resolution representation $\mathcal L_{\mathbf s}$ of its graph Laplacian using the transform in \eqref{eq:Ls} as a tensor,
i.e., ${|\mathbf{s}|\times N \times N}$ where $|\mathbf{s}|$ is the total number of scales. 
It maps $\mathcal L$ to a high-dimensional space of filtered graph edges with $|\mathbf{s}|$.  

{\bf FC Layer:} 
This component is a fully connected Deep Neural Network (DNN) classifier that takes the multi-resolution features of a graph as an input and predicts its class label. 
It has $L$ output units 
where $L$ is the number of available classes. 
The values $o_l$ computed at each output unit, when normalized, 
become the pseudo-probability of an input belong to a specific class.

MENET learns two sets of parameters: 1) scale 
$\mathbf{s}$ that
define the resolutions of graph edges,
and 2) weight $\mathbf{W}^{h}$ within the FC for prediction.
Notice that there is no pooling; 
while it is important to increase the efficiency of the algorithm, 
without an invertible method for pooling on edges, 
we want to keep {\em spatial interpretability} 
for neuroimaging applications. 
Also, the pooling in conventional 
methods are 
required for multi-scale analysis 
with fixed window size, but our framework inherits such 
behavior within the matrix transform.

\vspace{-8pt}
\subsection{Training MENET}
\vspace{-5pt}
Given a training set with $N_G$ number of individual graphs $G$ with corresponding labels $y$,
the learning process consists of feedforward and backpropagation steps.
In the feedforward propagation of MENET,
a graph $G$ is inputted to the framework as a graph Laplacian, and the probability that the $G$ belongs to a particular class 
is computed at the output.
Suppose we are given an input graph $G$ with $N$ vertices and a set of initial scales $\mathbf s$.
First, using the operation in \eqref{eq:Ls}, the input $G$ is transformed to $\mathcal{L}_s$. 
Since each $\mathcal{L}_s$ is represented as a matrix and there are $|\mathbf{s}|$ of them, 
$\mathcal L_s$ are combined to consist a feature map $M$ as a tensor: 
\begin{equation} 
\label{eq:mgr_graph_neural_network_imp}
M = \mathcal{L}_{s_0} \cup \mathcal{L}_{s_1} \cup \cdots \cup \mathcal{L}_{s_{|\mathbf{s}|}}
\end{equation}

Given $T$ hidden units in the hidden layers of the DNN module, at each $t$-th hidden unit,
the learned features in hidden units are linearly combined with corresponding weights in the first layer of DNN as
\begin{equation}
\label{eq:output_layer_b}
\footnotesize
z_{t} = \overset{|\mathbf{s}|,N,N}{\underset{s, p, q}{\sum}} w^{h}_{t,spq}m_{spq},
\end{equation}
where $m_{spq}$ represents a feature element in 
$M$ (i.e., an element at $(p,q)$ in $s$-th scale) and 
$w^{h}_{t,spq}$ denotes the weight on a connection between a hidden unit $m_{spq}$ to the $t$-th hidden unit of the following layer.
A non-linear activation function $\sigma()$ (e.g., sigmoid or rectified linear function) is applied on $z_t$ and is then fed to the next layer. 
At the output of the DNN, a soft-max function is used to get the final outcome $o_{l}$ for the $l$-th output unit 
as a pseudo probability from which a prediction is made as $\hat y = \arg\max_l o_l$. 

Once the output $o_l$ is obtained from the feedforward system above,
an error can be computed between the $o_l$ and the target value $y_{l}$, i.e., a label with one-hot-encoding for the input graph. 
We use focal loss \cite{lin2017focal} to measure the error: 
\begin{equation}
\footnotesize
 \label{eq:cost_ce}
 J(\mathbf{s}, \mathbf{W}^{h}) = -\frac{1}{N_{G}} \overset{N_{G}}{\underset{i=1}{\sum}} \overset{L}{\underset{l=1}{\sum}} -\alpha_{il}(1 - p_{il})^{\gamma}\log(p_{il})
\end{equation}
where $p_{il} = o_{il}$ 
if target value $y_{il} = 1$ at $l$-th output unit with $i$-th sample, otherwise $p_{il} = 1 - o_{il}$, $N_{G}$ and $L$ are the total number of graphs in a batch 
and the total number of available classes respectively, $\alpha$ and $\gamma$ are balanced variants. 
Our framework ``adaptively'' learns scale parameters $\mathbf{s}$ for novel graph representations 
and $\mathbf{W}^{h}$ in the FC layer (embedded in $o_l$) 
by primarily minimizing the classification error in \eqref{eq:cost_ce} via 
backpropagation. 
In traditional wavelet transform, the $\mathbf s$ is fixed to yield theoretical guarantees, however, 
we freely explore different aspects of $\mathbf s$ to find the optimal resolutions that yield the least loss. 

\noindent {\bf Regularization.} To avoid overfitting (especially with small sample size) and achieve 
desirable properties in the learned parameters,
we impose the following constraints to our model.
We first assume that
only a few edges in the graphs are 
highly associated with the variable of interest.
This is a natural assumption as changes due to a brain disorder do not manifest over the whole brain 
but sparsely appear in different ROIs. 
We therefore impose an $\ell_1$-norm constraint to the first layer of $\mathbf{W}^h$ which includes the fully connected weights.
We expect that this constraint will set many elements in the first layer of $\mathbf{W}^h$ to zeros and identify the edges that
are highly related to the prediction of labels.
Second, we expect $\mathbf s$ to be smooth with an $\ell_2$-norm constraint. This lets us obtain a smoothly transiting multi-resolution representation, 
i.e., avoid $\mathbf{s}$ from diverging. 

With these assumptions and \eqref{eq:cost_ce}, we minimize our final objective function
\begin{equation}\label{eq:new_obj}
\footnotesize
\tilde{J}(\mathbf{s}, \mathbf{W}^{h}) = J(\mathbf{s}, \mathbf{W}^{h}) + \frac{\theta_{1}}{N_{G}}|\mathbf{W}_{1}^{h}|_{1} + \frac{\theta_{2}}{N_{G}}||\mathbf{s}||_{2}
\end{equation}
where $\mathbf{W}_{1}^{h}$ represents the weights of the first layer of DNN module, $\theta_{1}$ and $\theta_{2}$ are the regularization parameters for $\ell_1$-norm and $\ell_2$-norm respectively. 

The MENET is trained based on the partial derivatives of the objective function \eqref{eq:new_obj} with respect to the trainable parameters, i.e., $\frac{\partial \tilde{J}(\mathbf{s}, \mathbf{W}^{h})}{\partial w^{h}_{t,spq}}$ and $\frac{\partial \tilde{J}(\mathbf{s}, \mathbf{W}^{h})}{\partial s}$. 
These parameters $\mathbf{W}^h$ and $\mathbf{s}$ are then updated using gradient descent with different learning rates $r_{W}$ and  $r_s$ respectively.

\vspace{-10pt}
\vspace{-5pt}
\section{Experiments}
\vspace{-10pt}
\label{sec:exp}
In this section, we demonstrate
experiments on two independent real datasets:
structural brain connectivity from Diffusion Tensor Images (DTI) in Alzheimer's Disease Neuroimaging Initiative (ADNI) and functional connectivity 
from resting-state functional magnetic resonance images (rs-fMRI) in ADHD-200.

\vspace{-10pt}
\subsection{Datasets}
\vspace{-5pt}

\noindent{\bf ADNI.} 
From the initiative, individual DTIs were processed by our in-house tractography pipeline to extract structural brain networks using Destrieux atlas \cite{destrieux2010automatic} with 148 ROIs. 
Each brain network is given as an adjacency matrix whose elements denote {\em number of neuron fiber tracts} connecting two different ROIs. 
The dataset included $N=506$ subjects 
and we merged control (CN) and Early Mild Cognitive Impairment (EMCI) groups as {\em Pre-clinical AD} group and combined Late Mild Cognitive Impairment (LMCI) and AD groups as {\em Prodromal AD} group to ensure sufficient sample size and compare their subtle differences. 

\noindent {\bf ADHD-200.} 
We adopted rs-fMRI data which were registered to Automated Anatomical Labeling (AAL) atlas with 116 ROIs \cite{tzourio2002automated}. 
We computed Pearson's correlation coefficients (without threshold) between 116 different ROIs to construct functional brain connectivity for each participant. 
Taking the samples without artifact and removing groups with few subjects, we ended up with total of $N$$=$$756$ samples 
labeled as 1) Typically Developing Children (TDC), 2) ADHD-Combined (ADHD-C), and 3) ADHD-Inattentive (ADHD-I). 
Our result may vary from \cite{10.3389/fnsys.2012.00069} as the experimental settings are different.

\vspace{-1.5em}
\begin{table}[!h]
\footnotesize
\caption{\footnotesize Demographics of ADNI and ADHD-200 Datasets}
\vspace{-1em}
\centering
\scalebox{0.75}{
\begin{tabular}{ |c||c|c||c|c|c| }
 \hline
 \multirow{2}{*}{Category} & \multicolumn{2}{c||}{\bf ADNI}  & \multicolumn{3}{c|}{\bf ADHD-200} \\ \cline{2-6}
  & Preclinical AD (CN,EMCI) & Prodromal AD (LMCI,AD)   & TDC & ADHD-C & ADHD-I\\
 \hline
 \# of Subjects  & 276 (109,167) & 171 (94,77)    & 487 & 159 & 110 \\
 \hline
Age (mean, std)  & 72.7(73.8,72.0),  6.9(5.8,7.5) & 74.2(72.6,76.1), 6.9(6.4,7.0)   & 12.2(3.3)  & 11.2(3.0)  &  12.0(2.6)\\
 \hline
 Gender (M/F) &  163/113 (57/52, 106/61) & 98/73 (51/43, 47/30)   &  258/229 & 130/29 &   85/25\\
\hline
\end{tabular}
\label{tab:demo_adni}}
\end{table}

\vspace{-10pt}
\subsection{Experimental Settings}
\vspace{-8pt}
\textbf{Evaluation measures.} We used 3-fold cross validation (CV) to evaluate our model and baselines with unbiased results. Evaluation measures were accuracy, precision, recall and F1-score averaged across the folds. 

\noindent\textbf{Parameters.} 
The kernel was defined as $k_s(x) = sx e^{-sx}$ to ensure that the $k()$ behaves as a band-pass filter 
(i.e., it achieves 0 at the origin), 
and the total number of scales to derive multi-resolution representation was $|\mathbf{s}|=5$. 
Weights were randomly initialized with Xavier initialization and the scales were uniformly selected between $[0.01, 2.5]$. 
The input to DL methods was the flattened $\mathcal{L}_{s}$ and the number of hidden units was set to 256. 
LeakyReLU activation function with negative slope 0.2 and batch normalization were applied to hidden units. 
$\ell_{1}$-norm was applied on the first layer of DNN with hyper-parameter $\theta_1$$=$$0.0001$ to achieve sparsity. $\ell_{2}$-norm was adopted on scales with hyper-parameter $\theta_2$$=$$0.001$. The learning rate of scale parameters was set to $r_{s} = 0.01$, and that of weight parameters was set to $r_{W} = 0.001$. 
The $\gamma$ for focal loss was set to 2. 

\noindent{\bf Baselines.} 
We used 
SVM with RBF kernel, Logistic regression (LR), 
DNN, graph kernel (Shortest Path (SP)) with SVM, and several state-of-the-art graph learning frameworks (i.e., graph2vec \cite{narayanan2017graph2vec} and DGCNN \cite{dgcnn}, and GCN \cite{kipf2016semi}) as baselines for comparisons. 
For SVMs, we used Principal Component Analysis (PCA) with 
the rank of 10 
to perform dimension reduction otherwise the models predicted all test samples 
as a single class with poor precision and recall. 
Node degree was used as node feature if a method required one. 

\vspace{-10pt}
\subsection{Structural Brain Connectivity Analysis on ADNI}
\label{sec:result_ADNI}
\vspace{-5pt}

\noindent {\bf Result.} 
Analysis on structural connectivity for Preclinical AD 
was performed. 
Binary classification task was designed to identify differences between the Preclinical ($N$$=$$276$) and Prodromal AD ($N$$=$$171$) groups instead of classifying four groups. 
As there is no effective treatment for AD, 
predicting the risk of developing dementia in the preclinical stage which is critical to enable disease-modifying interventions. 
Moreover, this task is particularly challenging since the physiological distinction between EMCI and LMCI, which 
lie along the decision boundary between Preclinical and Prodromal AD, is known to be especially subtle \cite{moradi2015machine,chincarini2011local}.

\begin{figure}[!t]
\centering
\includegraphics[width=0.49\textwidth, height=4.4cm]{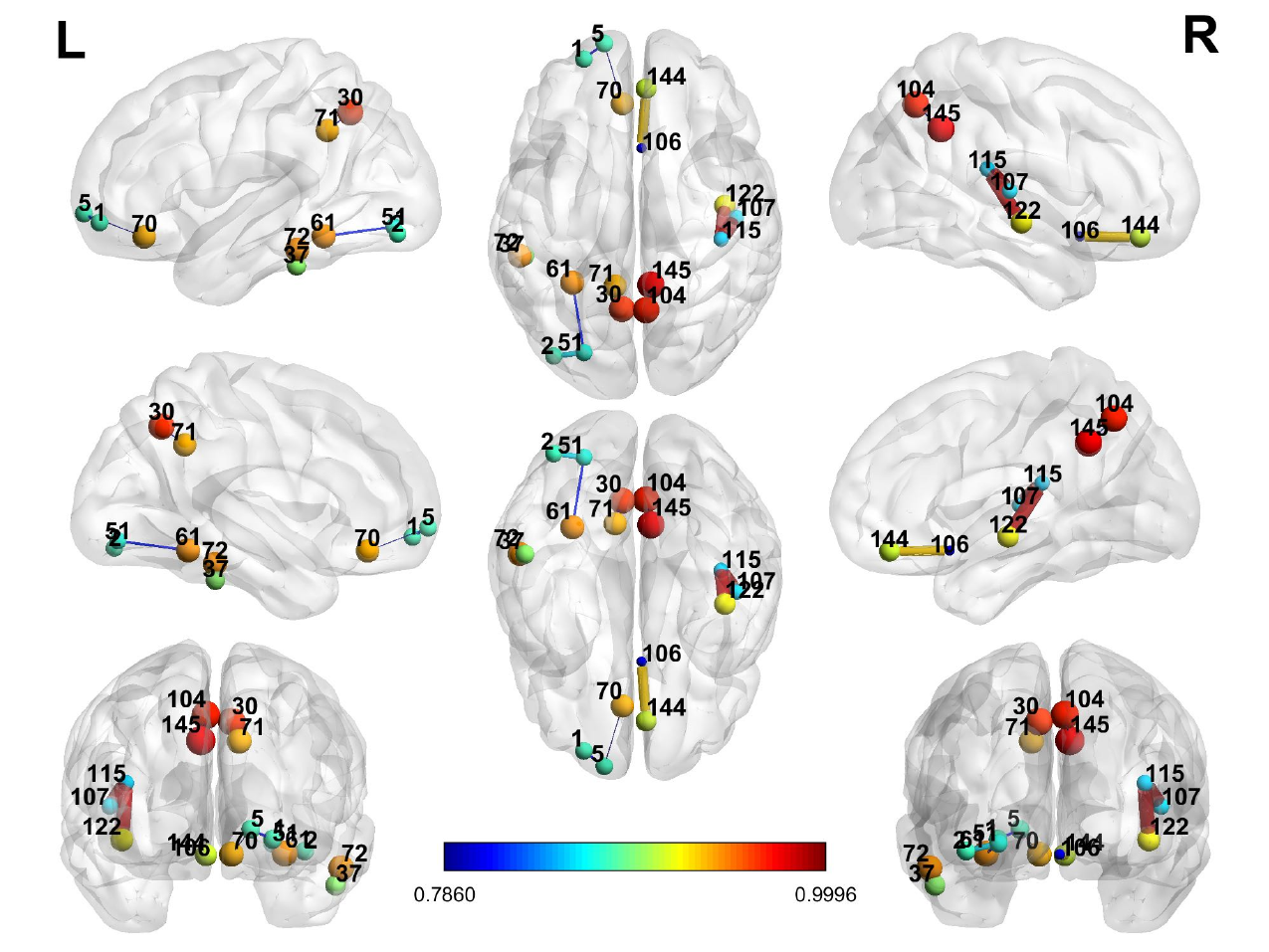}\hspace{-0.5em}
\includegraphics[width=0.49\textwidth, height=4.4cm]{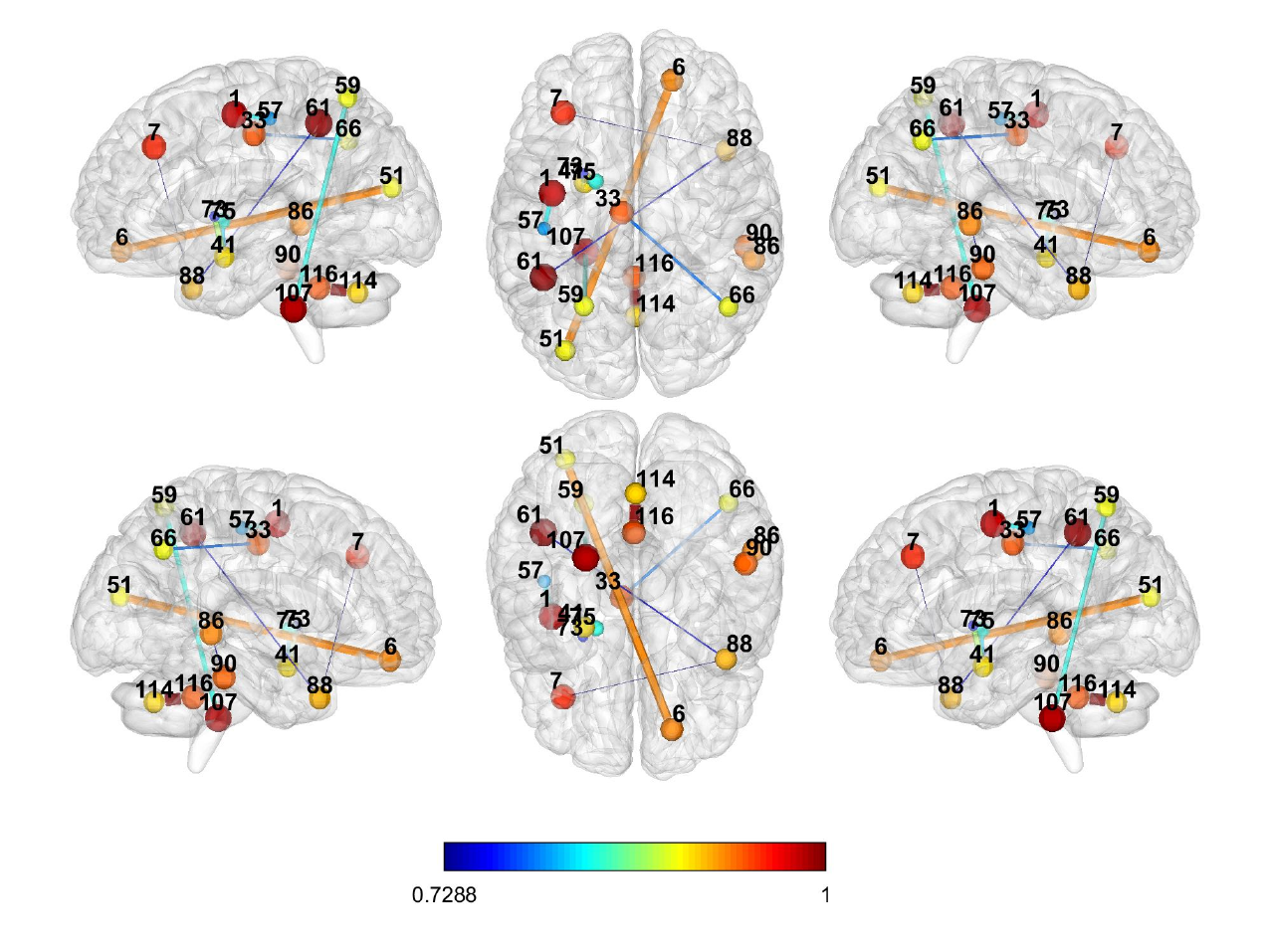}
\vspace{-1.4em}
\caption{\scriptsize  Top-10 Connectivities from ADNI (Left) and ADHD-200 (Right) Analyses. 
Edge thickness denotes average trained edge weight and node color denotes its degree.}
	\label{fig:adni_brain}
	\vspace{-2em}
\end{figure}

The classification results of all baselines and MENET across four evaluation measures (accuracy, precision, recall, and F1-score) averaged across the folds are shown in Table~\ref{tab:adni_result}.
We show that MENET outperforms \textit{all} other baselines. 
First, we see that SVM and graph kernel SVM achieved accuracy around $\sim60\%$, close to $61.7\%$ which is the random prediction would yield in this dataset, with poor precision and recall.
GNNs (i.e., DGCNN and GCN) did not perform well either; 
this may be due to both GCN and DGCNN emphasizing nodes instead of edges that better characterize these particular graphs.
While graph2vec with SVM showed higher performance than GNNs, 
its graph embedding is characterized by node and not edges, hence not interpretable at the connectivity level. 
Interestingly, directly applying LR on these $\mathcal L$ outperforms most of the previously mentioned baselines. This further demonstrates that the existing graph-based deep models may be \textit{sub-optimal} 
for the current task; instead, the key is to effectively operate on $\mathcal L$ as we observe next. 
\newline
\noindent{\bf Ablation Study ($\mathcal L$ vs. $\mathcal L_s$).} 
DNN directly applied on $\mathcal L$ immediately improves over LR. 
Although this gain is somewhat expected, we point out the significance of the multi-resolution graph transform $\mathcal L_s$: MENET, which is essentially DNN on $\mathcal L_s$ with the \textit{same} network classifier, shows significant improvements.
MENET improves over DNN 
(\textit{second best} model) 
in \textit{all} evaluations by 
$10.9\%$ in accuracy, $9.1\%$ in precision, $14.6\%$ in recall, and $14.0\%$ in F1-score.
\newline
\noindent {\bf Clinical validation.} To help clinical interpretation of our findings, we investigate 
the trained edge weights that are connected to the trained $\mathcal L_s$, 
and visualize them as a brain network in Fig.~\ref{fig:adni_brain}. 
Only top-10 connectivities with the 
largest average weights are shown for sparsity --- 
the most discriminative connectivities for 
Preclinical and Prodromal AD. 
The edge thickness 
and the node color 
respectively correspond to the edge weight and nodal degree of each ROI based on the trained weights. 
The list of the 10 connections is given along the figure that span across 17 ROIs (full ROI labels are 
in \cite{destrieux2010automatic}). 
We observed several temporal regions (i.e., inferior temporal gyrus (37), inferior temporal sulcus (72), superior temporal gyrus (107)) \cite{kim2012wavelet,galton2001differing}, precuneus (30, 104) \cite{karas2007precuneus}, subparietal regions (71, 145) \cite{choo2010posterior},
and many others, all of which are corroborated by various AD literature. 

\begin{table}[!b]
\vspace{-15pt}
\caption{\footnotesize List of Top-10 Significant Brain Connectivities from Fig. \ref{fig:adni_brain}. }
	\label{tab:roi_list}
	\vspace{-10pt}
\centering
\begin{minipage}{0.56\textwidth}
\centering
\scalebox{0.64}{
\begin{tabular}{|c|l|l|}
 \hline
 \multicolumn{3}{|c|}{\bf ADNI} \\ \hline
 {\bf Index}&{\bf Row (ROI label~\cite{destrieux2010automatic})} & {\bf Col (ROI label)}   \\ 
 \hline\hline
1 & 115 (rLat\_Fis-post) & 107 (rG\_temp\_sup-G\_T\_transv)    \\ 
 \hline
2 & 122 (rS\_circular\_insula\_inf) & 115 (rLat\_Fis-post)  \\ 
\hline
3 & 106 (rG\_subcallosal) & 144 (rS\_suborbital)   \\
\hline
4 & 51 (lS\_collat\_transv\_post)  & 2 (lG\_and\_S\_occipital\_inf) \\ 
\hline
5 & 72 (lS\_temporal\_inf) & 37 (lG\_temporal\_inf)   \\ 
\hline
6 & 51 (lS\_collat\_transv\_post) & 61 (lS\_oc-temp\_med\_and\_Lingual)    \\ 
 \hline
7 & 1 (lG\_and\_S\_frontomargin) & 5 (lG\_and\_S\_transv\_frontopol)  \\
\hline
8 & 30 (lG\_precuneus) & 71 (lS\_subparietal)   \\
\hline
9 & 145 (rS\_subparietal)  & 104 (rG\_precuneus) \\ 
\hline
10 & 5 (lG\_and\_S\_transv\_frontopol) & 70 (lS\_suborbital)   \\ 
\hline
\end{tabular}}
\end{minipage}
\begin{minipage}{0.44\textwidth}
\centering
\scalebox{0.64}{
\begin{tabular}{ |c|l|l|}
 \hline
 \multicolumn{3}{|c|}{\bf ADHD-200} \\ \hline
 \bf Index & \textbf{Row(ROI label~\cite{tzourio2002automated})} & \textbf{Col(ROI label)}   \\ 
 \hline\hline
 1 & 114 (Vermis\_8) &  116 (Vermis\_10)   \\ 
 \hline
 2 & 6 (Frontal\_Sup\_Orb\_R) & 51 (Occipital\_Mid\_L)   \\ 
\hline
 3 & 41 (Amygdala\_L) & 73 (Putamen\_L)  \\ 
\hline
 4 & 59 (Parietal\_Sup\_L) & 107 (Cerebelum\_10\_L)   \\ 
\hline
 5 & 1 (Precentral\_L) & 57 (Postcentral\_L)   \\ 
\hline
 6 & 41 (Amygdala\_L) & 75 (Pallidum\_L)   \\ 
\hline
 7 & 33 (Cingulum\_Mid\_L) & 66 (Angular\_R)   \\ 
\hline
 8 & 61 (Parietal\_Inf\_L) & 88 (Tem\_Pole\_Mid\_R)   \\ 
\hline
 9 & 86 (Temporal\_Mid\_R) & 90 (Temporal\_Inf\_R)   \\
 \hline
 10 & 7 (Frontal\_Mid\_L) & 88 (Tem\_Pole\_Mid\_R)   \\
\hline
\end{tabular}}
\end{minipage}
\end{table}

\vspace{-12pt}
\subsection{Functional Brain Connectivity Analysis on ADHD}
\label{sec:adhd}
\vspace{-7pt}

\noindent
{\bf Result.} 
Our results on multi-class classification on ADHD-200 are summarized in Table \ref{tab:adni_result}. 
We note that identifying the differences among these groups is innately challenging in ADHD-200 dataset consisting of adolescents with actively developing brains, inducing high variation. 
Further, compared to the previous AD experiment, this \textit{multi-class} classification is particularly difficult especially with the severe class imbalance ($64.4\%$ random test set prediction accuracy). 
Thus, it is crucial that a method benefits all four evaluations in this analysis.
In fact, throughout our experiments, we often observed undesirable behaviors from these algorithms biasing towards predicting all testing instances as TDC for maximal high accuracy while sacrificing precision and recall.
For instance, graph2vec+SVM achieved the third-highest accuracy with nearly the worst precision, recall and F1-score even with our best effort to alleviate this issue.
Therefore, all the models were carefully tuned to prevent such cases as much as possible.
Again, MENET achieved higher performance than all baseline models in \textit{all} evaluation metrics by achieving the highest average accuracy at $62.82\%$ while also improving precision, recall and F1-score.
DNN, DGCNN and GCN 
suffered from large standard deviations, indicating unstable learning across the CV folds.

\begin{table}[!t]
\caption{\small Classification Performances on ADNI and ADHD-200 Datasets.}
\vspace{-8pt}
\small
\centering
\scalebox{0.6}{
\begin{tabular}{ |c||c|c|c|c||c|c|c|c|}
 \hline
 \multirow{2}{*}{Model} & \multicolumn{4}{|c||}{\bf ADNI} &  \multicolumn{4}{|c|}{\bf ADHD-200} \\ \cline{2-5} \cline{6-9} 
  & Accuracy  & Precision & Recall & F1-score &  Accuracy  & Precision & Recall & F1-score \\ 
 \hline
SVM & $62.19\pm 5.86\%$  & $57.67\pm 8.90\%$ & $56.93\pm 7.17\%$  & $56.21\pm 8.49\%$ &  $52.06\pm 4.39\%$  & $31.95\pm 3.11\%$ & $32.56\pm 2.28\%$  & $31.75\pm 1.98\%$ \\
 \hline
LR  & $74.27\pm 2.70\%$  & $74.47\pm 1.91\%$ & $69.60\pm 4.15\%$ & $70.08\pm 4.63\%$ &  $45.29\pm 8.36\%$  & $27.18\pm 4.56\%$ & $28.74\pm 3.50\%$ & $27.21\pm 3.53\%$ \\
\hline
SP-SVM & $58.61\pm 1.14\%$  & $42.02\pm 4.58\%$ & $48.13\pm 1.14\%$  & $39.77\pm 1.50\%$ &  $51.13\pm 4.25\%$  & $31.84\pm 1.06\%$ & $32.36\pm 1.00\%$  & $31.63\pm 0.68\%$ \\
\hline
DNN & $78.65\pm 0.92\%$  & $80.24\pm 1.15\%$ & $74.49\pm 2.35\%$  &  $75.45\pm 2.13\%$ &  $56.04\pm 7.17\%$  & $41.94\pm 9.23\%$ & $34.63\pm 0.96\%$  &  $31.91\pm 3.45\%$ \\
\hline
graph2vec  & $74.83\pm 5.93\%$  & $75.06\pm 5.56\%$ & $71.89\pm 7.25\%$ & $71.69\pm 7.64\%$ &  $61.62\pm 1.79\%$  & $21.36\pm 0.08\%$ & $31.82\pm 0.93\%$ & $25.56\pm 0.35\%$ \\
\hline
DGCNN  & $65.77\pm 2.90\%$  & $73.20\pm 11.48\%$ & $62.20\pm 3.31\%$ &  $61.89\pm 4.03\%$ &  $62.27\pm 3.12\%$  & $37.47\pm 13.34\%$ & $35.20\pm 1.95\%$ &  $29.94\pm 4.06\%$ \\ \hline
GCN  & $67.79\pm 3.71 \%$  & $67.67\pm 7.30\%$ & $62.34\pm 3.33\%$ &  $62.21\pm 3.44\%$ &  $59.76\pm  6.23\%$  & $39.68\pm 10.91\%$ & $33.66\pm 0.37\%$ &  $29.49\pm 3.08\%$ \\
\hline
{\bf MENET} &  $\bf 87.27 \pm 3.71\%$  &  $ \bf 87.56 \pm 2.90\%$ &  $ \bf 85.39 \pm 5.06\%$  & $ \bf 86.02 \pm 4.41\%$ &   $\mathbf{62.82\pm 1.60\%}$  &  $\mathbf{44.95\pm 4.46\%}$ &  $\mathbf{35.48\pm 1.82\%}$  & $\mathbf{32.41\pm 3.38\%}$ \\
\hline
\end{tabular}
\label{tab:adni_result}}
\vspace{-17pt}
\end{table}

\noindent{\bf Clinical Validation.} To help clinical interpretation of our findings, we similarly 
derived the top-10 functional connectivities that are associated with the classification of ADHD stages. The top-10 connectivities with the highest average weights are shown in Fig.~\ref{fig:adni_brain}, whose edge thickness and node color correspond to the edge weight and degree of each ROI respectively. The list of the 10 connections, that span across 18 ROIs, is given along the figure (full ROI names in \cite{tzourio2002automated}). We observed several ROIs in temporal regions (mid-temporal pole (86, 88), inferior temporal cortex (90)) \cite{rubia2007temporal}, frontal regions (frontal superior orbital (6), mid-frontal (7)) \cite{murias2006functional},
and Amygdala (41) \cite{marsh2008reduced}; these results along with other identified ROIs are already well documented in many ADHD literature. 

\vspace{-15pt}
\subsection{Discussions on Convergence of Scales}
\vspace{-8pt}

\begin{wrapfigure}[9]{r}{0.55\textwidth}
\vspace{-25pt}
\centerline{
    \includegraphics[width=0.98\linewidth]{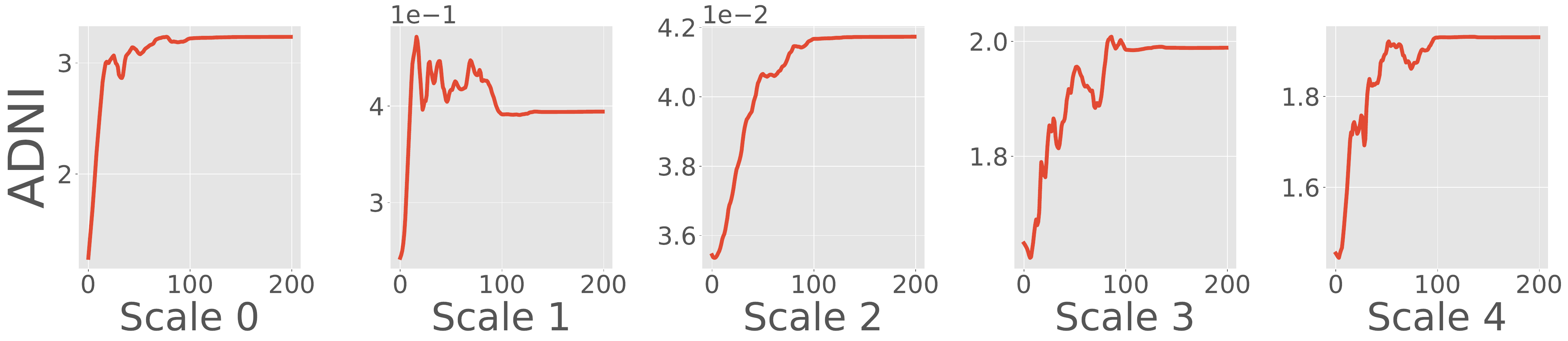}
    }
\centerline{
    \includegraphics[width=0.98\linewidth]{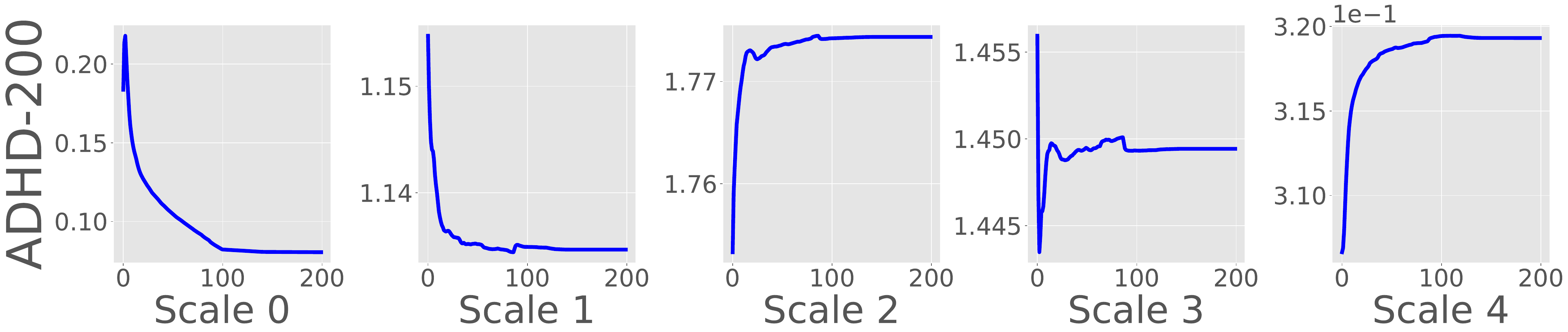}
    }
\vspace{-10pt}
  \caption{\scriptsize Convergence of scales w.r.t. training epoch. Top: ADNI, Bottom: ADHD-200. 
	}\label{fig:scale_convergence}
	\vspace{-1.8em}
\end{wrapfigure}
Fig.~\ref{fig:scale_convergence} shows the convergence of scales for AD (top) and ADHD (bottom) experiments. 
The scale parameters of our model converge very fast;
the low-pass filter (i.e., Scale 0) of the ADNI experiment tends to increase while ADHD tends to decrease in the experiments. One possible reason is that 
the ADHD classification requires the model to involve more local context information as the problem is more difficult than the AD experiment.  

\vspace{-10pt}
\vspace{-5pt}
\section{Conclusion}
\vspace{-10pt}
We developed a novel graph transform-based CNN framework designed to perform  classification tasks with a population of registered graphs. 
The transform derives multi-resolution representations of a graph matrix, i.e., edges, 
that serve as effective features 
suited to perform classification on graphs. 
Using a parametric kernel, our framework, i.e., MENET, can train well with relatively small sample size and was validated with extensive experiments on two independent connectivity datasets, yielding clinically sound results on AD and ADHD supported by existing literature. 
We believe that MENET has significant potential to domains with graph data practically challenged by small sample sizes. 
\vspace{-12pt}

\newline

\vspace{3pt}
\noindent {\textbf{Acknowledgments}}. This research was supported by NSF IIS CRII 1948510, NSF IIS 2008602, NIH R01 AG059312, IITP-2020-2015-0-00742, and IITP-2019-0-01906 funded by MSIT (AI Graduate School Program at POSTECH).

\vspace{-12pt}
%
%
{\tiny
\bibliographystyle{splncs04}
\bibliography{ipmi_2021.bib}}

\end{document}